\documentclass[letterpaper, 10 pt, journal, twoside]{IEEEtran}  

\usepackage{graphicx} 
\graphicspath{{figures/}}

\usepackage{amsthm}
\usepackage{amsmath} 
\usepackage{amssymb}  
\usepackage{mathtools}
\usepackage{ulem} 
\usepackage{multirow}
\usepackage{verbatim}
\usepackage{soul}
\usepackage{algorithm}
\usepackage{algpseudocode}

\usepackage{color}
\usepackage{xcolor}
\usepackage{latexsym}
\usepackage{multicol}
\usepackage{booktabs}
\usepackage{adjustbox}
\usepackage[table]{xcolor}
\usepackage{booktabs}
\usepackage{makecell}

\usepackage{enumitem}
\usepackage[font=small,labelfont=bf]{caption}
\usepackage{lipsum}
\usepackage{subcaption}
\usepackage[hidelinks]{hyperref}



\renewcommand{\thefigure}{\arabic{figure} }

\IEEEoverridecommandlockouts                              

\begin{document}

\title{\LARGE \bf
Reflex: Enabling Fast and Predictive Vision-Language-Action Models for Reaction-Critical Manipulation
}

\author{Yuxuan Chen$^{1}$, Wanruo Zhang$^{1}$, and Xiao Li$^{1}$

\thanks{$^{1}$ Shanghai Jiao Tong University
         {\tt\footnotesize chen\_yuxuan@sjtu.edu.cn}}}%



\maketitle

\begin{abstract}
Vision-Language-Action (VLA) models have recently achieved promising performance in robotic manipulation. However, existing benchmarks mainly evaluate generalization on static manipulation tasks and largely overlook dynamic interaction scenarios. To address this gap, we present \textit{ReflexBench}, a benchmark for reaction-critical manipulation. ReflexBench contains six dynamic tasks and introduces an evaluation framework that decouples simulator stepping from robot control while supporting configurable latency under synchronous and asynchronous inference. Building upon ReflexBench, we propose \textit{ReflexVLA}, an efficient VLA model designed for reaction-critical manipulation without large-scale robot-data pretraining. ReflexVLA enhances temporal reasoning through latent future prediction and multi-frame temporal fusion within the vision backbone, while reducing deployment latency through batched visual encoding and CUDA Graph replay. Experiments show that ReflexVLA consistently improves dynamic manipulation performance while maintaining competitive accuracy on standard static manipulation benchmarks, and real-world experiments further demonstrate its effectiveness under practical deployment conditions. Project website: \href{https://reflexvla.github.io/}{reflexvla.github.io}
\end{abstract}

\section{Introduction}
Recent advances in Vision-Language-Action (VLA) models have significantly expanded the capabilities of robotic manipulation by unifying perception, language understanding, and action generation within a single framework. Building upon the success of large-scale vision-language models, a growing body of work demonstrates that robot policies can be learned from diverse multimodal data and generalized across tasks through natural language instructions. Representative approaches \cite{brohan2023rt} \cite{kim2024openvla} \cite{black2410pi0} show that scaling model capacity and training data leads to substantial improvements in task diversity, instruction following, and cross-domain generalization. These developments establish VLA models as a promising foundation for general-purpose robot control and motivate further exploration toward increasingly capable embodied agents.

Despite these advances, current VLA models remain challenged in manipulation scenarios that require both rapid reaction and reasoning about future states. Many existing approaches primarily condition actions on the current observation, and their inference latency introduces a delay between perception and execution. This delay becomes particularly consequential in dynamic environments where robots must respond to moving objects, intercept targets at precise times, or manipulate objects whose trajectories affect task success.

 \begin{figure}[t]
    \centering
    \includegraphics[width=0.95\linewidth]{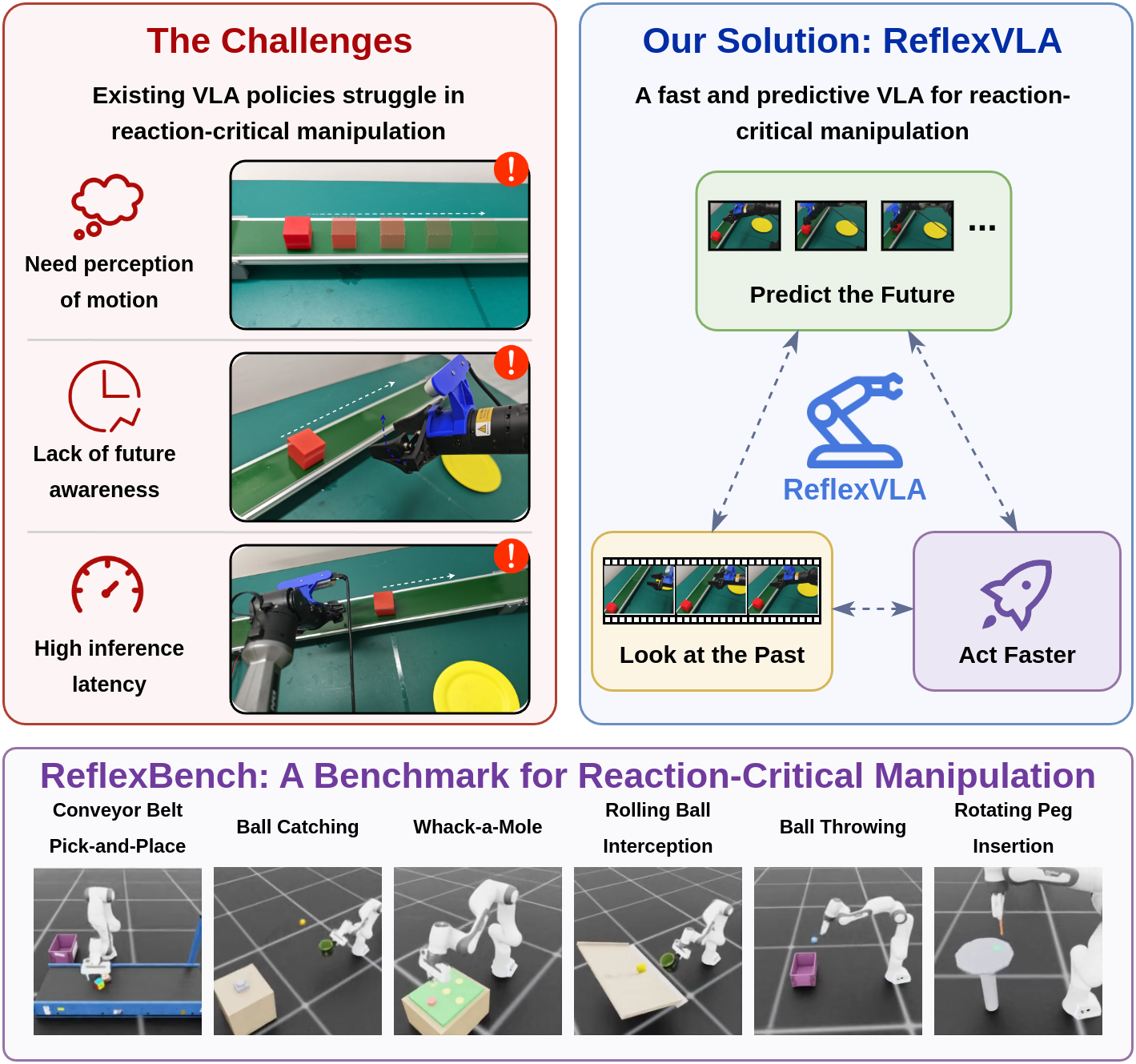}
    \caption{\textbf{Overview of our work.} We introduce \textbf{ReflexBench}, a benchmark of six reaction-critical manipulation tasks. Building upon it, we propose \textbf{ReflexVLA}, a fast and predictive VLA that achieves strong performance with low latency.}
    \label{fig:f1}
\end{figure}

To address these limitations, recent research explores several complementary directions. One line of work introduces specialized benchmarks that focus on dynamic manipulation scenarios \cite{xie2026dynamicvla} \cite{fang2026towards} \cite{zhang2026overcoming}. In parallel, a growing number of studies investigate how to equip VLA models with capabilities that are better suited for future prediction \cite{xie2026dynamicvla} \cite{lu2026faster}. Meanwhile, considerable effort is devoted to improving the efficiency of VLA inference through architectural optimization, model compression, and system-level acceleration techniques, with the goal of reducing latency and enabling deployment in real-time robotic applications \cite{shukor2025smolvla} \cite{ma2025running}. However, these research directions are largely explored in isolation. Existing dynamic manipulation benchmarks primarily serve as evaluation platforms without providing a unified framework for studying both anticipation and efficiency, while methods that enhance predictive reasoning often introduce additional computational overhead. Conversely, approaches that prioritize inference acceleration generally focus on efficiency alone and do not explicitly address the need for future-aware decision making. As a result, the challenge of simultaneously achieving strong anticipatory capability and fast inference remains insufficiently explored in current VLA research.

In this paper, we first present \textbf{ReflexBench}, a benchmark designed to evaluate VLA policies in manipulation tasks that require both efficient execution and anticipation of future environmental dynamics. ReflexBench consists of six simulation-based tasks that capture representative challenges in reaction-critical manipulation. To better reflect practical deployment conditions, the benchmark explicitly incorporates latency effects and aligns the evaluation process as closely as possible with real-world execution delays. Building upon ReflexBench, we further investigate the key factors that enable VLA policies to succeed in such tasks and propose \textbf{ReflexVLA}, a VLA designed for fast and predictive robot control. ReflexVLA is based on three core components. First, it introduces latent prediction of future visual observations, which encourages the policy to reason about near-future environmental states. Second, it leverages multiple historical observations as input, providing richer temporal context for decision making. Third, it incorporates inference latency optimization to reduce the delay between perception and action execution. Extensive experiments demonstrate that these design choices contribute to improved performance, and their combination leads to a more effective solution for dynamic manipulation scenarios that demand both anticipation and efficiency. To summarize our contributions, we

\begin{itemize}
    \item introduce ReflexBench, a benchmark consisting of six manipulation tasks that require both rapid reaction and future-aware decision making;
    \item propose ReflexVLA, a fast and predictive VLA framework that combines future latent prediction, multi-frame history modeling, and inference latency optimization;
    \item conduct extensive experiments in both simulation and real-world environments, demonstrating the effectiveness of the proposed designs.
\end{itemize}

\section{Related Work}\label{sec:literature review}

\subsection{Benchmarks for Dynamic Manipulation}

Recent benchmarks have begun evaluating VLA models in dynamic environments. \cite{xie2026dynamicvla} focuses on object motion and perception-execution latency, while \cite{fang2026towards} studies spatiotemporal reasoning across 35 dynamic manipulation tasks. \cite{zhang2026overcoming} further reveals that current VLA models remain highly sensitive to environmental dynamics. In contrast, ReflexBench explicitly models inference latency under both synchronous and asynchronous execution, providing a more faithful evaluation of latency-sensitive robotic manipulation.

\subsection{Efficient Vision-Language-Action Models}
As VLA models continue to grow in scale, improving their efficiency becomes increasingly important for real-world deployment. One line of work aims to reduce model size and training cost while maintaining competitive performance, like SmolVLA~\cite{shukor2025smolvla} and VLA-Adapter~\cite{wang2026vla}. Another direction focuses on improving inference efficiency at the system and architecture levels. FASTER~\cite{lu2026faster} revisits the design of flow-based VLAs and explores mechanisms for real-time action generation. VLASH~\cite{tang2025vlash} further improves responsiveness through future-state-aware asynchronous inference, reducing the impact of policy latency during execution. In the context of dynamic manipulation, DynamicVLA~\cite{xie2026dynamicvla} also considers the challenges introduced by perception-execution delays and investigates VLA architectures that are better suited for dynamic object interaction. Despite these advances, existing efficient VLA methods primarily focus on reducing computational cost or improving execution frequency. Comparatively less attention is devoted to understanding how inference latency interacts with future anticipation in dynamic manipulation.


\section{ReflexBench}
Recent robotic manipulation benchmarks evaluate VLA models across diverse tasks and environments \cite{liu2023libero} \cite{tao2024maniskill3} \cite{yu2020meta} \cite{li2023behavior}, but largely overlook reaction-critical scenarios. Furthermore, most simulation benchmarks pause the environment during policy inference, ignoring inference latency and the resulting perception-execution delays. Consequently, they fail to accurately assess VLA performance in dynamic real-world settings.

To address these limitations, we introduce ReflexBench, a benchmark specifically designed for reaction-critical manipulation tasks. ReflexBench explicitly incorporates realistic latency effects into the evaluation process and focuses on scenarios that require both future-aware decision making and efficient policy execution.

\subsection{Task List}
ReflexBench consists of 6 manipulation tasks that require a policy to infer quickly under dynamic environmental conditions, as shown in Fig. \ref{fig:f1}. Concretely, these tasks are:
\begin{itemize}
    \item \textbf{Conveyor Belt Pick-and-Place.} A cube and a distractor object move along a conveyor belt. The robot is required to pick up the cube and place it into a nearby bin.
    \item \textbf{Ball Catching.} The robot must catch an incoming ball whose position changes rapidly over time.
    \item \textbf{Whack-a-Mole.} Targets appear at different locations for a short duration and disappear if not reached in time.
    \item \textbf{Rolling Ball Interception.} A ball rolls down a slope, and the robot must catch it at the edge of the slope to prevent it from falling off.
    \item \textbf{Ball Throwing.} The robot throws a ball toward a bin.
    \item \textbf{Rotating Peg Insertion.} The robot inserts a peg into a continuously rotating socket.
\end{itemize}


\subsection{Data Collection}

\begin{figure*}[htb]
    \centering
    \includegraphics[width=1.7\columnwidth]{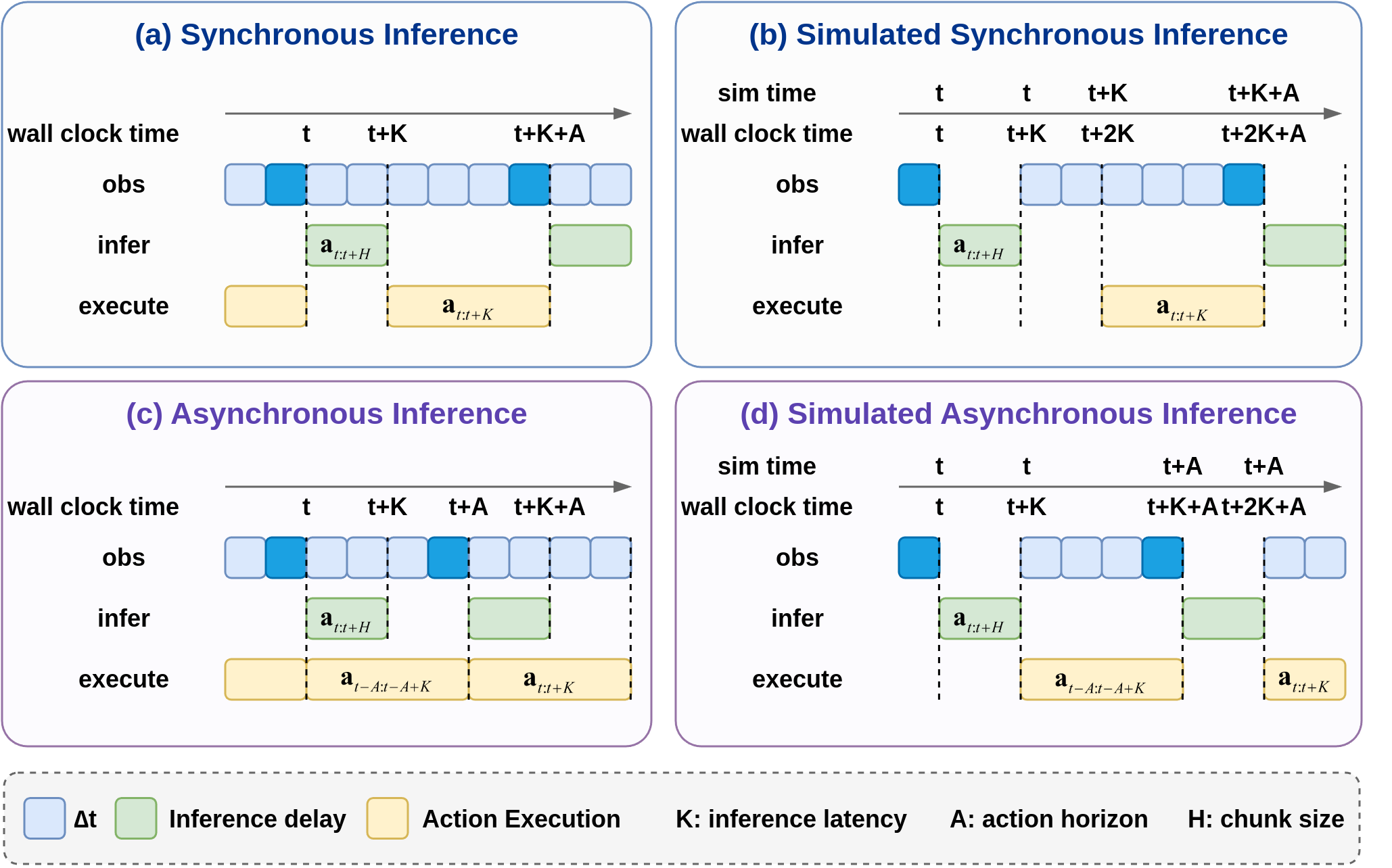}
    \caption{\textbf{Illustration of Four Inference Mechanisms.} ReflexBench models the effect of policy inference latency by explicitly simulating the temporal gap between perception and action execution while allowing the latency to be adjusted in a controllable manner.}
    \label{fig:infer}
\end{figure*}

Collecting demonstrations for dynamic manipulation tasks is substantially more challenging than for conventional static manipulation benchmarks. In static environments, expert trajectories can often be generated through straightforward scripted policies or teleoperation. In contrast, the tasks in ReflexBench involve moving objects and time-dependent interactions, where successful execution requires continuous adaptation to environmental dynamics.

We employ a planning-based data collection pipeline that decomposes each task into sequential phases with distinct subgoals. For example, in Rotating Peg Insertion, the robot performs pre-alignment, target tracking, and timed insertion. To handle dynamic objects, the planner predicts future trajectories from real-time position and velocity, allowing actions to be planned toward anticipated future states. When planning alone cannot handle complex interaction dynamics, we train a task-specific reinforcement learning policy to generate demonstration trajectories.

\subsection{Latency-Aware Evaluation}
A key feature of ReflexBench is its latency-aware evaluation protocol. Unlike existing simulation benchmarks that pause the environment during policy inference, ReflexBench decouples simulation from robot control, allowing environmental dynamics to evolve independently of policy computation and enabling more realistic evaluation.

ReflexBench explicitly models inference latency under two common deployment paradigms: synchronous and asynchronous inference. Synchronous inference blocks execution until policy inference completes, whereas asynchronous inference overlaps action execution with policy computation, improving responsiveness at the cost of slightly stale actions.

To faithfully reproduce latency effects in simulation, we introduce a latency blocking mechanism. Fig.~\ref{fig:infer} illustrates the two inference paradigms and their corresponding simulation protocols. Under synchronous inference, the simulator pauses to collect the current observation and query the policy. After inference, the simulator resumes, the robot remains idle for the specified latency, and then executes the predicted action chunk, faithfully mimicking inference-blocking deployment. Under asynchronous inference, the policy infers a new action chunk from the current observation while the robot executes the action chunk generated in the previous cycle. The newly inferred action becomes available in the next control period.

The latency value can be specified manually, which effectively changes the duration of the blocking window without modifying the policy or hardware configuration. This enables systematic analysis of how different latency levels affect task performance under controlled conditions. In addition, ReflexBench supports latency settings derived from actual policy execution. We first measure the real-world inference latency of a policy and then convert it into the corresponding simulation delay using the Real-Time Factor (RTF), which is defined as
\begin{equation}
\mathrm{RTF} = \frac{t_{\mathrm{sim}}}{t_{\mathrm{wall}}},
\end{equation}
where $t_{\mathrm{sim}}$ denotes the elapsed simulation time and $t_{\mathrm{wall}}$ denotes the corresponding wall-clock time required to execute the simulation. The RTF characterizes the relative speed of the simulator with respect to real time. Given a measured policy inference latency in the real world, we compute the equivalent delay in simulation according to the simulator's RTF and inject the corresponding latency into the evaluation process.

\section{ReflexVLA}

\begin{figure*}[htb]
    \centering
    \includegraphics[width=1.9\columnwidth]{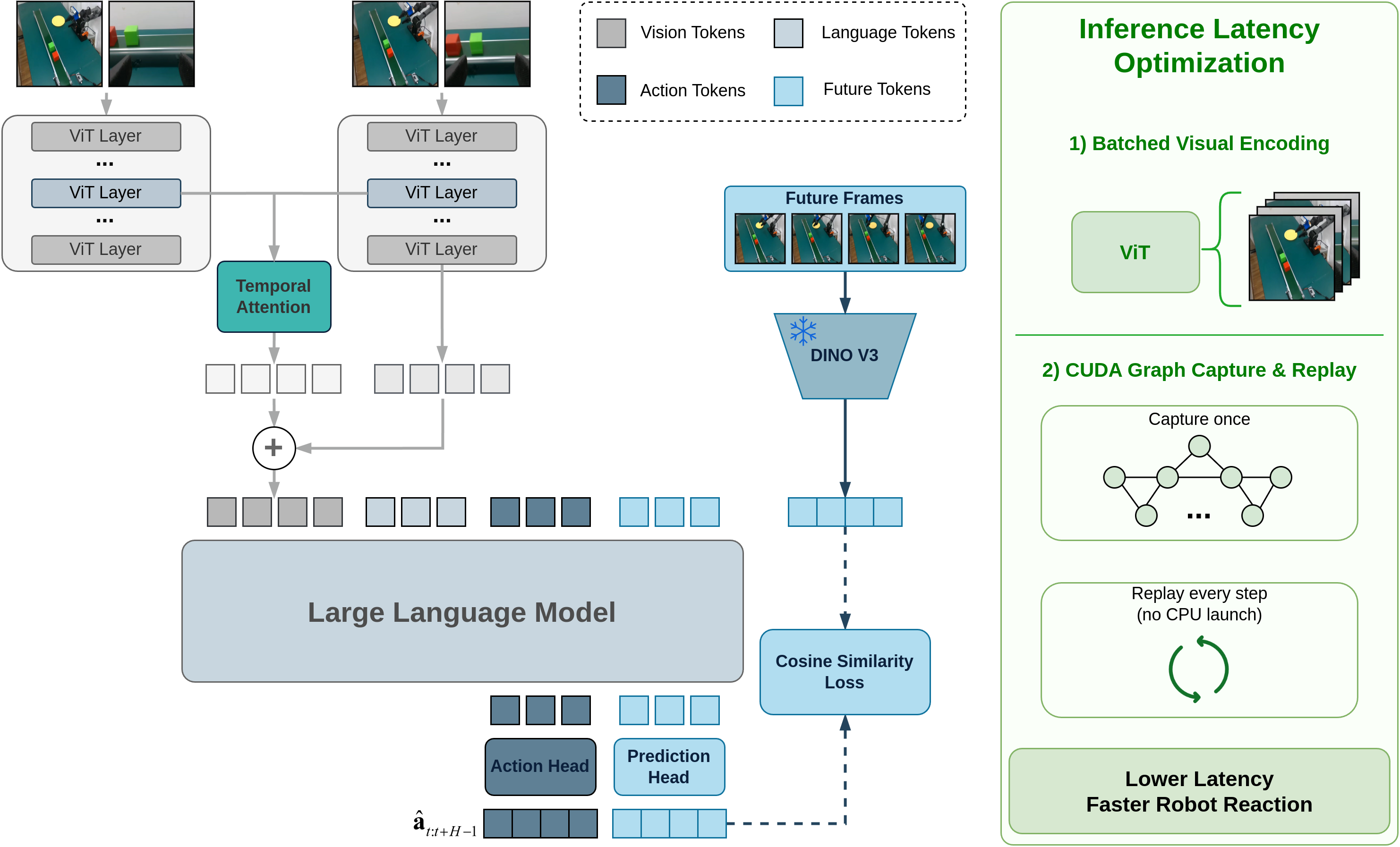}
    \caption{\textbf{Overview of ReflexVLA.} ReflexVLA is an efficient vision-language-action model for reaction-critical robotic manipulation, integrating latent future prediction, multi-frame temporal fusion, and inference latency optimization to enable fast and anticipatory decision-making.}
    \label{fig:architecture}
\end{figure*}

ReflexVLA is built on top of a compact vision-language-action architecture \cite{wang2026vla}. The visual encoder is a fused ViT backbone composed of DINOv2 \cite{oquab2024dinov2} and SigLIP \cite{zhai2023sigmoid} at $224 \times 224$ resolution. The language backbone is a Qwen2.5-0.5B \cite{qwen2.5} causal transformer. Given a language instruction and multi-view RGB observations, the model inserts projected visual tokens after the beginning-of-sequence token and uses learnable action queries to produce an action chunk through a continuous regression head. As shown in Fig. \ref{fig:architecture}, ReflexVLA introduces three key designs: latent future prediction for anticipatory representation learning, multi-frame temporal fusion for short-horizon motion awareness, and inference-time computation optimization for responsive deployment.

\subsection{Latent Future Prediction}

Dynamic manipulation often requires anticipating future states rather than reacting to the current observation. To this end, ReflexVLA incorporates an auxiliary latent future prediction objective to learn predictive scene representations.

Directly predicting future images is computationally expensive and often forces the model to focus on low-level appearance details that are not directly relevant to control. Instead, we perform future prediction in a semantic latent space extracted by a frozen visual encoder. For each training sample at time step $t$, the dataset provides a future observation horizon $\{\mathbf{o}_{t+i}\}_{i=1}^{H}$. We encode each future observation using a frozen DINOv3 \cite{simeoni2025dinov3} model:
\begin{equation}
    \mathbf{y}_{t+i}
    =
    \phi_{\mathrm{DINOv3}}(\mathbf{o}_{t+i}),
    \qquad
    \mathbf{y}_{t+i}\in\mathbb{R}^{1024}.
\end{equation}

To enable future prediction, we append $H$ learnable future tokens to the multimodal input sequence. The prediction horizon $H$ is chosen to be identical to the chunk size used by action prediction, such that each future token corresponds to one future control step within the predicted action chunk. This design aligns future state prediction with the temporal structure of action generation, allowing the model to learn scene dynamics that are directly relevant to downstream control. And each future step is represented by only a single prediction token, making this design significantly efficient. Since the number of prediction tokens is equal to the action chunk size, the additional sequence length remains small in practice. Consequently, the computational and memory overhead introduced by future prediction is negligible relative to the overall cost of the VLA backbone.


After the transformer forward pass, the hidden representation associated with the $i$-th future token is projected into the DINOv3 feature space:
\begin{equation}
    \hat{\mathbf{y}}_{t+i}
    =
    f_{\mathrm{pred}}
    \left(
    \mathbf{h}^{\mathrm{future}}_{i}
    \right),
    \qquad
    i=1,\ldots,H.
\end{equation}

The predicted feature is trained to match the corresponding future visual representation using a masked cosine similarity loss:
\begin{equation}
    \mathcal{L}_{\mathrm{future}}
    =
    \frac{
    \sum_{i=1}^{H}
    m_i
    \left[
    1-
    \cos\left(
    \hat{\mathbf{y}}_{t+i},
    \mathbf{y}_{t+i}
    \right)
    \right]
    }{
    \sum_{i=1}^{H} m_i
    },
\end{equation}
where $m_i$ indicates whether the corresponding future observation is valid.

The future prediction loss is jointly optimized with the action prediction objective:
\begin{equation}
    \mathcal{L}
    =
    \mathcal{L}_{\mathrm{act}}
    +
    \lambda_{\mathrm{future}}
    \mathcal{L}_{\mathrm{future}},
\end{equation}
where $\mathcal{L}_{\mathrm{act}}$ denotes the action learning loss and $\lambda_{\mathrm{future}}$ controls the contribution of future prediction. By requiring the policy to anticipate future visual states during training, ReflexVLA learns representations that capture the temporal evolution of dynamic scenes.

\subsection{Multi-Frame Temporal Fusion}
Since motion dynamics cannot be captured from a single observation, the policy requires historical visual context. A straightforward solution is to concatenate multi-frame visual tokens, but this introduces $VTP$ tokens for $V$ views and $T$ frames, with $P$ visual tokens per image, resulting in high computational costs due to quadratic attention complexity. Moreover, it ignores explicit temporal structures and relies on the language model to learn cross-frame interactions implicitly.

To address these limitations, ReflexVLA performs temporal fusion within the vision backbone and exposes only the fused representation of the current frame to the language model. Given $V$ camera views and $T$ historical observations per view, the vision encoder first extracts patch-level features for all images:
\begin{equation}
    \mathbf{X}^{\mathrm{mid}},
    \mathbf{X}^{\mathrm{final}}
    \in
    \mathbb{R}^{B \times V \times T \times P \times D},
\end{equation}
where $\mathbf{X}^{\mathrm{mid}}$ and $\mathbf{X}^{\mathrm{final}}$ denote intermediate-layer and final-layer visual features, respectively.

We perform temporal fusion on intermediate features, which retain richer local appearance and motion cues than deeper, more semantic representations.

For each camera view $v$ and patch location $p$, features from the same spatial position across the $T$ historical frames form a temporal trajectory
$\mathbf{x}^{\mathrm{mid}}_{v,p,1:T}$. The features are first normalized and projected into a lower-dimensional space:
\begin{equation}
    \mathbf{z}_{v,p,1:T}
    =
    \mathrm{Down}
    \left(
    \mathrm{LN}
    \left(
    \mathbf{x}^{\mathrm{mid}}_{v,p,1:T}
    \right)
    \right)
    +
    \mathbf{e}_{1:T},
\end{equation}
where $\mathbf{e}_{1:T}$ denotes temporal positional embeddings and $\mathrm{Down}(\cdot)$ reduces the feature dimension to improve efficiency.

A causal temporal attention layer is then applied:
\begin{equation}
    \Delta \mathbf{x}_{v,p,t}
    =
    \mathrm{Up}
    \left(
    \mathrm{MHA}_{\mathrm{causal}}
    \left(
    \mathbf{z}_{v,p,1:T}
    \right)_T
    \right).
\end{equation}
The causal mask prevents information leakage from future observations and ensures consistency with online deployment. The resulting feature summarizes the relevant motion history for the current patch. 


The resulting temporal representation is integrated into the current-frame visual feature:
\begin{equation}
    \tilde{\mathbf{x}}_{v,p,t}
    =
    \mathbf{x}^{\mathrm{final}}_{v,p,t}
    +
    \Delta \mathbf{x}_{v,p,t}.
\end{equation}
All fused current-frame tokens are then flattened as
\begin{equation}
    \tilde{\mathbf{X}}_t
    \in
    \mathbb{R}^{B \times VP \times D},
\end{equation}
projected into the language model embedding space, and inserted into the multimodal sequence for action prediction.

An important advantage of this design is that the number of visual tokens consumed by the language model remains identical to that of the single-frame setting. Consequently, ReflexVLA can exploit short-term motion information without introducing the substantial language-model-side overhead associated with multi-frame images.

\subsection{Inference Latency Optimization}

Inference latency is a critical factor in dynamic manipulation, where even strong policies can fail if actions are generated too slowly. To improve real-world responsiveness, ReflexVLA reduces end-to-end inference latency by optimizing two common sources of overhead in VLA systems: visual processing and GPU execution.

The first optimization targets visual encoding. Since ReflexVLA consumes multiple camera views and historical observations, a naive implementation would independently process each image through the vision backbone, resulting in $V \times T$ separate forward passes. Such repeated execution introduces unnecessary framework overhead and leads to suboptimal GPU utilization. Instead, ReflexVLA batches all view-frame images into a single visual encoder invocation.

A single batched forward pass then produces visual features for all observations simultaneously, after which the resulting features are reshaped into $\mathbb{R}^{B \times V \times T \times P \times D}$
for temporal fusion. This formulation naturally aligns with the multi-frame architecture of ReflexVLA while significantly reducing per-frame execution overhead.

The second optimization focuses on GPU execution efficiency. During deployment, the computation graph of ReflexVLA remains fixed, including visual encoding, temporal fusion, multimodal projection, language model inference, and action prediction \cite{ma2025running}. Rather than repeatedly launching individual GPU kernels for every control cycle, we capture the complete inference pipeline as a CUDA Graph:
\begin{equation}
    \hat{\mathbf{a}}_{t:t+H-1}
    =
    \mathcal{G}
    \left(
    \mathbf{x}_{\mathrm{text}},
    \mathbf{x}_{\mathrm{vision}},
    \mathbf{x}_{\mathrm{state}}
    \right),
\end{equation}
where $\mathcal{G}$ denotes the captured CUDA Graph and $\hat{\mathbf{a}}_{t:t+H-1}$ represents the predicted action chunk. During execution, new observations are copied into pre-allocated buffers and the graph is replayed directly, eliminating repeated kernel scheduling and runtime dispatch overhead.

Notably, these optimizations do not modify the policy architecture or learning objective. Instead, they reduce deployment overhead while preserving the original multimodal reasoning process. The system-level optimizations substantially improve execution efficiency and make ReflexVLA more suitable for latency-sensitive robotic manipulation tasks.

\section{Experiments}
\label{sec: experiments and results}

In our experiments, we focus on the following questions:
\begin{itemize}
    \item Q1: How do different factors influence the success rate of reaction-critical manipulation tasks?
    \item Q2: Does ReflexVLA achieve better performance on ReflexBench than existing VLA baselines?
    \item Q3: Does ReflexVLA still achieve competitive performance on standard static manipulation tasks?
    \item Q4: What is the contribution of each design component to the overall performance of ReflexVLA?
    \item Q5: How well does ReflexVLA perform in real-world manipulation tasks?
\end{itemize}

\subsection{Setup}

\textbf{Baselines.} We compare ReflexVLA with a diverse set of representative VLA models. Specifically, we include the lightweight models VLA-Adapter \cite{wang2026vla} and SmolVLA \cite{shukor2025smolvla}, the dynamic manipulation methods DynamicVLA \cite{xie2026dynamicvla} and PUMA \cite{fang2026towards}, and the large-scale VLA models OpenVLA-OFT \cite{kim2025fine} and $\pi_{0.5}$ \cite{pmlr-v305-black25a}.

\textbf{Metrics.} We evaluate task performance using \textit{task success rate} and report \textit{model size} as an indicator of deployment cost. We additionally measure end-to-end \textit{inference latency} in the latency ablation. 

\textbf{Implementation Details.} Unless otherwise specified, all VLA policies are trained as a single policy on the same dataset, consisting of 200 demonstration episodes for each of the 6 tasks. ReflexVLA is trained with 2 consecutive temporal observations as visual input and a latent future prediction weight of $\lambda_{\mathrm{future}}=0.05$. All policies use asynchronous inference with an action chunk size of 8 and an action horizon of 2, and are evaluated on a single NVIDIA RTX 5880 Ada GPU. Each success rate is computed over 150 evaluation episodes per task and reported as the mean and standard deviation across three runs with different random seeds.

\subsection{Main Results}

 \begin{figure}[t]
    \centering
    \includegraphics[width=0.95\linewidth]{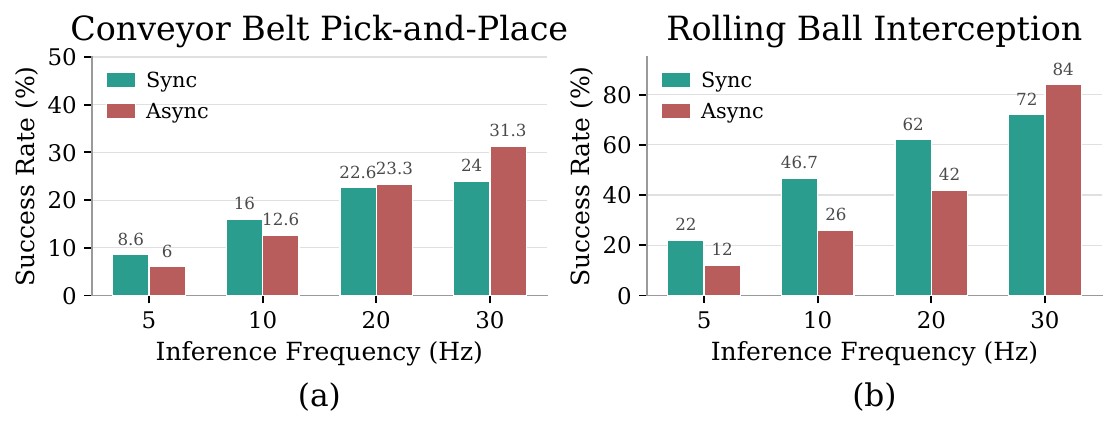}
    \caption{Success rates of SmolVLA under synchronous and asynchronous inference across different inference frequencies on two tasks. Chunk size is set to 8.}
    \label{fig:latency}
\end{figure}

 \begin{figure}[t]
    \centering
    \includegraphics[width=0.95\linewidth]{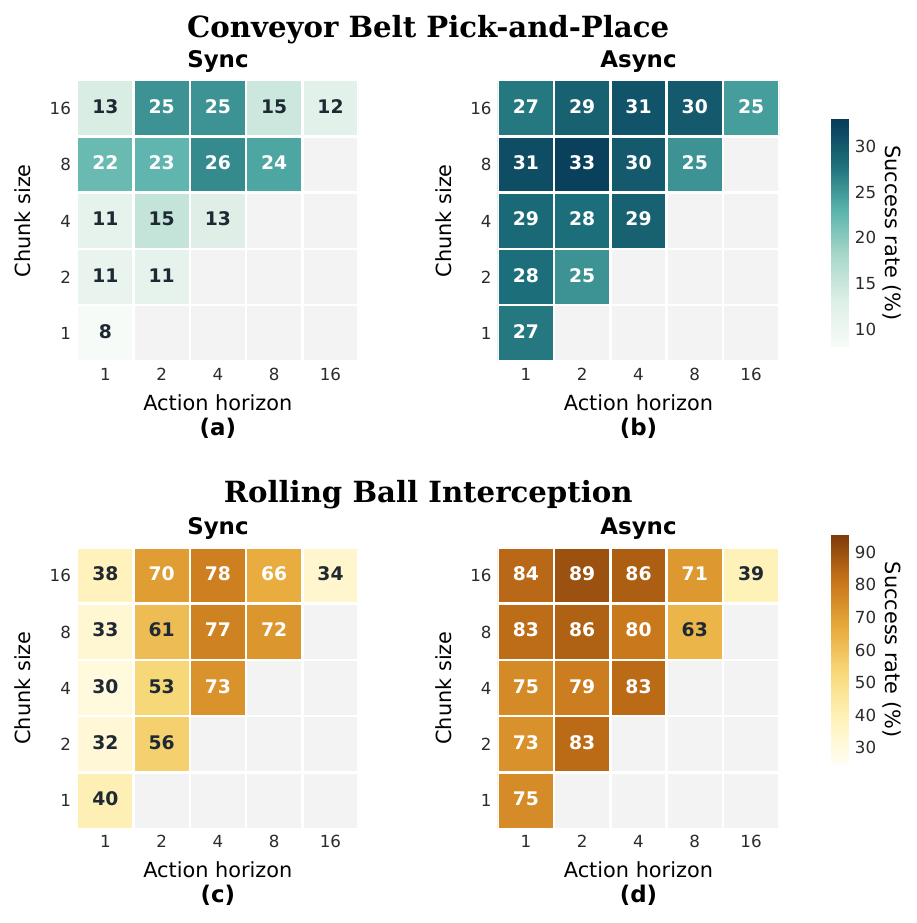}
    \caption{Success rate matrices of SmolVLA under different chunk sizes and action horizons on two tasks. Chunk size is the policy's training hyperparameter, while action horizon is the number of action steps executed during inference. The inference frequency is fixed at~30 Hz.}
    \label{fig:chunk}
\end{figure}

\begin{table*}[t]
\centering
\caption{Performance comparison on ReflexBench.
Results are reported as mean success rate (\%) $\pm$ standard deviation over three runs.}
\label{tab:main_results}

\small
\setlength{\tabcolsep}{5pt}
\renewcommand{\arraystretch}{1.15}

\begin{tabular}{lcccccccc}
\toprule
\textbf{Model} &
\textbf{Params} &
\makecell[c]{\textbf{Conveyor Belt}\\\textbf{Pick-and-Place}} &
\makecell[c]{\textbf{Ball}\\\textbf{Catching}} &
\makecell[c]{\textbf{Whack-a-}\\\textbf{Mole}} &
\makecell[c]{\textbf{Rolling Ball}\\\textbf{Interception}} &
\makecell[c]{\textbf{Ball}\\\textbf{Throwing}} &
\makecell[c]{\textbf{Rotating Peg}\\\textbf{Insertion}} &
\textbf{Avg.} \\
\midrule

OpenVLA-OFT \cite{kim2025fine}
& 7B
& 58.0$\pm$1.2
& 5.3$\pm$1.2
& \textbf{100.0$\pm$0.0}
& 41.4$\pm$4.2
& 10.0$\pm$3.3
& 1.3$\pm$0.7
& 36.0 \\

$\pi_{0.5}$ \cite{pmlr-v305-black25a}
& 4B
& 39.1$\pm$1.0
& 6.0$\pm$0.7
& 98.9$\pm$0.3
& 36.8$\pm$1.8
& 34.0$\pm$2.7
& 6.7$\pm$1.4
& 36.9 \\

PUMA \cite{fang2026towards}
& 4B
& 67.4$\pm$1.7
& 4.0$\pm$0.7
& \textbf{100.0$\pm$0.0}
& \textbf{85.1$\pm$3.4}
& 33.8$\pm$1.0
& 11.1$\pm$1.7
& 50.2 \\

DynamicVLA \cite{xie2026dynamicvla}
& 0.5B
& 30.6$\pm$2.4
& 4.4$\pm$1.0
& 90.2$\pm$1.7
& 45.8$\pm$1.7
& \textbf{36.2$\pm$4.7}
& 16.7$\pm$4.2
& 37.3 \\

SmolVLA \cite{shukor2025smolvla}
& 0.5B
& 19.3$\pm$2.0
& 2.9$\pm$0.8
& \textbf{100.0$\pm$0.0}
& 70.2$\pm$2.0
& 27.1$\pm$1.0
& 10.0$\pm$4.6
& 38.3 \\

VLA-Adapter \cite{wang2026vla} (Baseline)
& 1B
& 36.8$\pm$2.4
& 6.0$\pm$2.0
& 68.4$\pm$3.9
& 23.1$\pm$7.7
& 29.1$\pm$5.2
& \textbf{18.4$\pm$2.5}
& 30.3 \\

\midrule

\rowcolor{gray!10}
\textbf{ReflexVLA (Ours)}
& \textbf{1B}
& \textbf{73.8$\pm$4.1}
& \textbf{7.3$\pm$0.7}
& \textbf{100.0$\pm$0.0}
& 77.1$\pm$7.1
& 31.7$\pm$1.0
& 12.4$\pm$1.5
& \textbf{50.4} \\

\bottomrule
\end{tabular}
\end{table*}

\begin{figure*}[htb]
    \centering
    \includegraphics[width=1.8\columnwidth]{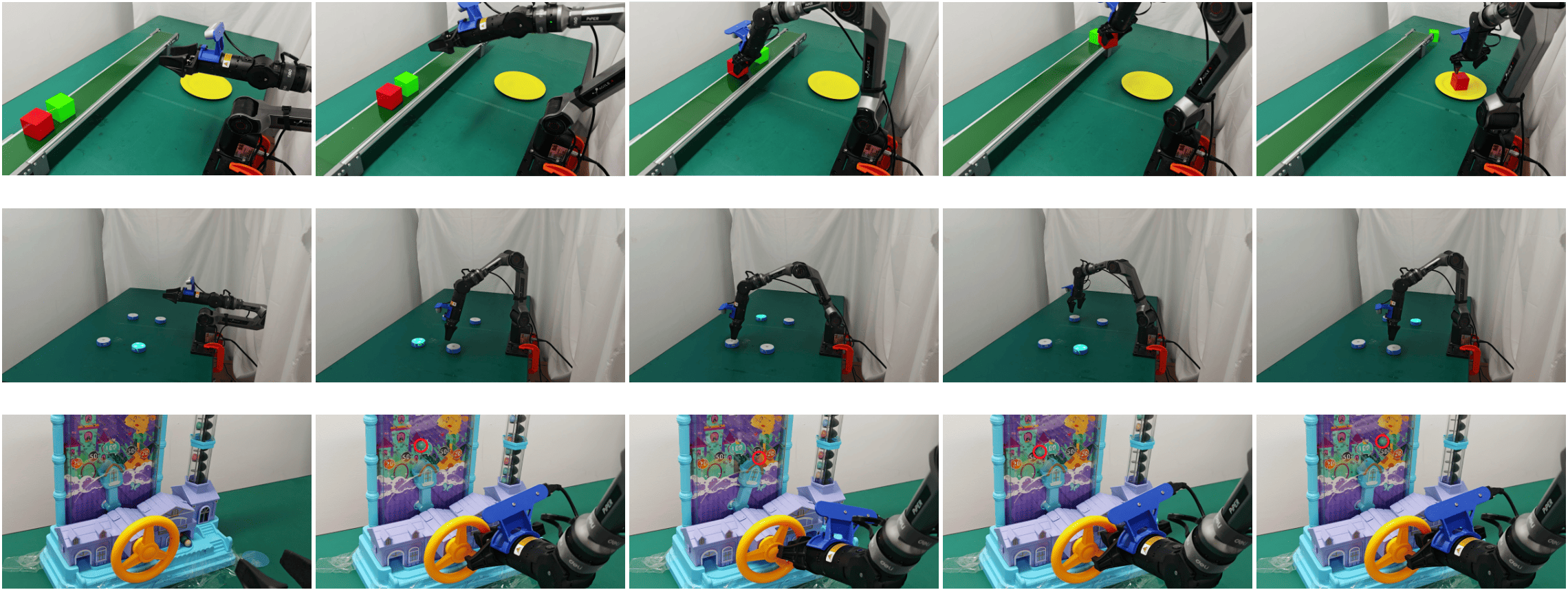}
    \caption{\textbf{Real-world rollout examples.} From top to bottom: Conveyor Belt Pick-and-Place, PressButtons, and CatchBalls.}
    \label{fig:real_world}
\end{figure*}

\textbf{The best performance is achieved by combining a larger chunk size with a shorter action horizon under high-frequency asynchronous inference (Q1).} Before evaluating different VLA policies, we first investigate how the inference paradigm, inference frequency, action chunk size, and action horizon affect policy performance on reaction-critical manipulation tasks. Here, the chunk size specifies the number of actions predicted per chunk, while the action horizon determines how many actions are executed before the next inference. To enable efficient exploration, we conduct this study using the lightweight SmolVLA model on two representative ReflexBench tasks, namely Conveyor Belt Pick-and-Place and Rolling Ball Interception.

Fig. \ref{fig:latency} compares synchronous and asynchronous inference across different inference frequencies. At low frequencies, asynchronous inference suffers from greater observation-action mismatch, which is particularly detrimental in rapidly changing environments. As the inference frequency increases, this mismatch is reduced, enabling asynchronous inference to increasingly benefit from continuous action execution and eventually outperform synchronous inference.

Fig. \ref{fig:chunk} studies the effect of chunk size and action horizon at a fixed inference frequency of 30~Hz. Larger chunk sizes with shorter action horizons under asynchronous inference generally achieve better performance, although the optimal chunk size varies slightly across tasks. We therefore use asynchronous inference with a chunk size of 8 and an action horizon of 2 for all subsequent experiments on ReflexBench to ensure fair comparisons.

\textbf{ReflexVLA achieves competitive average success rates on ReflexBench while maintaining a relatively compact model size (Q2).} Table~\ref{tab:main_results} demonstrates that ReflexVLA consistently achieves competitive performance across the six tasks in ReflexBench and obtains the highest overall average success rate among all evaluated methods. Specifically, ReflexVLA achieves an average success rate of $50.4 \%$, outperforming all existing baselines while using only a 1B-parameter model. Compared with its backbone model, VLA-Adapter, ReflexVLA improves the average success rate from $30.3 \%$ to $50.4 \%$, demonstrating the effectiveness of the proposed designs. Notably, ReflexVLA surpasses substantially larger general VLA models, including OpenVLA-OFT (7B) and $\pi_{0.5}$ (4B), and achieves comparable overall performance to the dynamic-oriented PUMA while requiring only one-quarter of its model size.

\begin{table}[t]
\centering
\caption{Results on LIBERO. Accuracy is reported in (\%).}
\label{tab:libero}

\footnotesize
\setlength{\tabcolsep}{5pt}
\renewcommand{\arraystretch}{1.1}

\begin{tabular}{lccccc}
\toprule
\textbf{Model} &
\textbf{Spatial} &
\textbf{Object} &
\textbf{Goal} &
\textbf{Long} &
\textbf{Avg.} \\
\midrule

OpenVLA-OFT \cite{kim2025fine}
& 97.6
& 98.4
& 97.9
& 94.5
& 97.1 \\

$\pi_{0.5}$ \cite{pmlr-v305-black25a}
& \textbf{98.8}
& 98.2
& \textbf{98.0}
& 92.4
& 96.9 \\

VLA-Adapter \cite{wang2026vla}
& 97.8
& \textbf{99.2}
& 97.2
& \textbf{95.0}
& \textbf{97.3} \\

\midrule

\textbf{ReflexVLA (Ours)}
& 98.2
& \textbf{99.2}
& \textbf{98.0}
& 93.6
& 97.2 \\

\bottomrule
\end{tabular}
\end{table}

\textbf{ReflexVLA maintains competitive performance on standard static manipulation benchmarks while being primarily designed for dynamic manipulation scenarios (Q3).} Table~\ref{tab:libero} reports the results on the LIBERO benchmark \cite{liu2023libero}. Although ReflexVLA is specifically designed to improve reaction-critical dynamic manipulation, it achieves an average success rate of $97.2 \%$, which is comparable to the best-performing VLAs. Compared with its backbone model, VLA-Adapter, ReflexVLA exhibits only a marginal difference in overall performance, indicating that the proposed future prediction and temporal modeling modules do not compromise the model's capability on conventional static manipulation tasks. These results demonstrate that the improvements introduced by ReflexVLA enhance dynamic manipulation performance while preserving strong generalization to standard robotic manipulation benchmarks.

\subsection{Ablation Study}

\definecolor{MyGreen}{RGB}{0,135,81}
\definecolor{MyRed}{RGB}{203,67,53}

\newcommand{\gain}[1]{{\scriptsize\textcolor{MyGreen}{(+#1)}}}
\newcommand{\drop}[1]{{\scriptsize\textcolor{MyRed}{(-#1)}}}

\begin{table}[t]
\centering
\small
\setlength{\tabcolsep}{5pt}
\caption{Progressive ablation study of ReflexVLA.}
\label{tab:ablation}

\begin{tabular}{llcc}
\toprule
\textbf{Method} & \textbf{Variant} & \textbf{SR (\%)} & \textbf{Lat. (ms)} \\
\midrule

Baseline
& --
& 36.8
& 81.522 \\

\midrule

\multirow{2}{*}{+ Future Pred.}
& Trainable
& 4.9 \drop{31.9}
& 82.508 \\
& Frozen
& 62.8 \gain{26.0}
& 82.508 \\

\midrule

\multirow{3}{*}{+ Temporal Fusion}
& Cross Attn.
& 66.1 \gain{29.3}
& 127.335 \\
& MHA
& 68.2 \gain{31.4}
& 125.824 \\
& MHA (Middle)
& 71.7 \gain{34.9}
& 125.107 \\

\midrule

+ Latency Opt.
& \makecell[l]{Batch Enc.\\+ CUDA Graph}
& \textbf{73.8} \gain{37.0}
& \textbf{64.991} \\

\bottomrule
\end{tabular}
\end{table}

\textbf{Latent future prediction enhances anticipatory reasoning, multi-frame temporal fusion improves motion understanding, and inference latency optimization reduces deployment delay, together leading to higher success rate with lower inference latency (Q4).} Table~\ref{tab:ablation} presents a progressive ablation study of ReflexVLA, where each proposed component is introduced incrementally starting from the baseline model.

We first investigate the latent future prediction module. Directly training the model to predict future visual representations with a trainable visual target results in a substantial performance degradation, indicating that jointly optimizing the representation space and the prediction objective leads to unstable supervision. In contrast, using features extracted from a frozen DINOv3 encoder provides a stable semantic prediction target and improves the success rate from $36.8\%$ to $62.8\%$ with almost no additional inference overhead.

Building upon this model, we further evaluate three temporal fusion strategies. Notably, performing temporal fusion on intermediate visual features instead of the final-layer representations yields the best performance of $71.7\%$. This observation suggests that intermediate features preserve richer motion-related information, making them more suitable for modeling temporal dynamics than highly semantic final-layer features.

Finally, we incorporate the proposed inference latency optimization, including batched visual encoding and CUDA Graph replay. These optimizations further improve the success rate to $73.8\%$ while simultaneously reducing the inference latency from $125.1$~ms to $65.0$~ms. The improved success rate demonstrates that reducing deployment latency is particularly important for reaction-critical manipulation tasks.

\subsection{Real-World Experiments} 

\begin{table}[t]
\centering
\small
\caption{\textbf{Quantitative results of the real-world experiments.} Conveyor Belt reports the number of successful trials out of 20 attempts. PressButtons reports the average number of buttons pressed within 30 seconds, and CatchBalls reports the average number of balls caught out of 10 balls. Each policy is evaluated 20 times on every task.}
\label{tab:real_world}
\begin{tabular}{lccc}
\toprule
Model & Conveyor & PressButtons & CatchBalls \\
\midrule
SmolVLA \cite{shukor2025smolvla} & 2/20 & 0.9 & 3.8 \\
PUMA \cite{fang2026towards}    & 13/20 & 20.8 & 5.4 \\
Ours    & \textbf{16/20} & \textbf{22.5} & \textbf{6.7} \\
\bottomrule
\end{tabular}
\end{table}

\textbf{Real-World Setup.} We validate ReflexVLA through real-world experiments on an AgileX Piper robotic arm. Three representative tasks are chosen: Conveyor Belt Pick-and-Place, PressButtons, and CatchBalls. Performance is measured by task-specific metrics: success rate for Conveyor Belt Pick-and-Place, correctly pressed buttons within 30 seconds for PressButtons, and successful catches out of 10 attempts for CatchBalls. We compare ReflexVLA with two representative baselines, SmolVLA \cite{shukor2025smolvla} and PUMA \cite{fang2026towards}, with all policies trained using teleoperation demonstrations.

\textbf{ReflexVLA demonstrates strong real-world manipulation performance (Q5).} Table~\ref{tab:real_world} presents the quantitative results, and Figure~\ref{fig:real_world} provides representative execution examples. Compared with SmolVLA, ReflexVLA consistently achieves higher performance across all three tasks, indicating that its temporal modeling and latency-aware designs are critical to these tasks. Despite using a much smaller model than PUMA, ReflexVLA achieves comparable overall performance on these dynamic manipulation tasks, suggesting that modeling future scene evolution together with efficient temporal perception is an effective approach for real-world reactive manipulation.

\subsection{Limitations} 
Our work has several limitations that warrant future investigation. First, although ReflexVLA introduces latent future prediction and multi-frame temporal fusion, these components are incorporated only during fine-tuning rather than through large-scale pretraining, which may limit their full potential. Second, our study focuses on two widely adopted inference paradigms, naive synchronous and asynchronous inference. More advanced inference mechanisms, such as RTC \cite{NEURIPS2025_300ccb21}, are not explored in this work and may provide additional improvements in latency-sensitive manipulation scenarios.

\section{Conclusion}
\label{sec:conclusion}
In this work, we present ReflexBench, a benchmark for latency-aware dynamic manipulation. ReflexBench contains six representative dynamic tasks and introduces a latency-aware evaluation framework that decouples simulator stepping from robot control. We also present ReflexVLA, a latency-aware vision-language-action framework for dynamic robotic manipulation. By integrating latent future prediction, multi-frame temporal fusion, and inference optimization, ReflexVLA enables robots to better anticipate future states while achieving efficient real-time execution. Experiments on ReflexBench demonstrate that ReflexVLA achieves competitive performance among existing VLA baselines while maintaining the efficiency of a lightweight 1B parameter model. Furthermore, ReflexVLA remains competitive on the standard static manipulation benchmark LIBERO. Real-world experiments on three reaction-critical manipulation tasks further validate the effectiveness and practicality of our approach.

\bibliographystyle{IEEEtran}
\bibliography{references.bib}

\end{document}